\documentclass[conference]{IEEEtran}

\usepackage[pdftex]{graphicx}
\usepackage[ruled,vlined,linesnumbered]{algorithm2e}
\usepackage{array}
\usepackage{cite}
\usepackage{url}

\begin{document}

\newenvironment{conditions}
  {\par\vspace{\abovedisplayskip}\noindent\begin{tabular}{>{$}l<{$} @{${}={}$} l}}
  {\end{tabular}\par\vspace{\belowdisplayskip}}

\title{Text Compression for Sentiment Analysis via Evolutionary Algorithms}

\author{\IEEEauthorblockN{Emmanuel Dufourq}
\IEEEauthorblockA{University of Cape Town\\ African Institute for Mathematical Sciences \\
Email: edufourq@gmail.com}
\and
\IEEEauthorblockN{Bruce A. Bassett}
\IEEEauthorblockA{University of Cape Town\\ African Institute for Mathematical Sciences \\ South African Astronomical Observatory \\
Email: bruce.a.bassett@gmail.com}
}

%

\maketitle

\begin{abstract}
Can textual data be compressed intelligently without losing accuracy in evaluating sentiment? In this study, we propose a novel evolutionary compression algorithm, PARSEC (PARts-of-Speech for sEntiment Compression), which makes use of Parts-of-Speech tags to compress text in a way that sacrifices minimal classification accuracy when used in conjunction with sentiment analysis algorithms. An analysis of PARSEC with eight commercial and non-commercial sentiment analysis algorithms on twelve English sentiment data sets reveals that accurate compression is possible with (0\%, 1.3\%, 3.3\%) loss in sentiment classification accuracy for (20\%, 50\%, 75\%) data compression with PARSEC using LingPipe, the most accurate of the sentiment algorithms. Other sentiment analysis algorithms are more severely affected by compression. We conclude that significant compression of text data is possible for sentiment analysis depending on the accuracy demands of the specific application and the specific sentiment analysis algorithm used. 
\end{abstract}

\begin{IEEEkeywords}
sentiment analysis, evolutionary algorithm, compression
\end{IEEEkeywords}

\IEEEpeerreviewmaketitle


\section{Introduction and Rationale}\label{Introduction}

Sentiment analysis is the automated detection of emotions and attitudes towards a particular subject, event or entity, and has received an increasing amount of interest since the year 2000 \cite{Pang:2008:Opinion,liu:2015:sentiment}. Sentiment analysis has been applied to many problem domains; for instance, determining sentiments of consumers towards products, or mining social media to gain an understanding of the public's opinion on matters such as corruption \cite{Pang:2008:Opinion,liu:2015:sentiment,Ravi:2015:ASurvey}. 

In this study, we propose a novel evolutionary algorithm (EA) which is tasked with compressing text, whereby the objective is to retain the correct sentiment classification of the compressed text. To achieve this, the EA makes use of POS (Parts of Speech) to determine which POS can be removed from the text without hindering the classification performance. EAs \cite{Eiben:2003:IntroTo,Yu:2012:Introduction} -- inspired by nature -- make use of a population of individuals which are evolved over several iterations to solve optimisation problems.

To illustrate the idea, consider the sentence ``I went home yesterday, opened the door and was immediately faced with a terrible shock". The key sentiment-encoding phrase is ``terrible shock". We thus want to propose an algorithm that can learn to evolve rules that can effectively do this compression. The remaining words in the sentence do not contain any useful information in determining the sentiment expressed, and thus the rules can discard those words.

Feng et al. \cite{Feng:2010:AChinese} proposed a Chinese text compression method with the objective of preserving the opinions expressed in the text to retain the sentiment, and to ensure that the compressed sentence is grammatically correct. The authors propose a score function which takes into consideration three primary functions; namely a word significance, a linguistic and an opinion scoring function. The word significance function makes use of the word frequency properties to allocate a score to each word. The scores were obtained by processing a corpus containing 30,000 documents. The linguistic score was obtained using a \textit{n-gram} probabilistic function which made use of Google search results. The opinion score made use of an opinion lexicon to allocate a value to each word based on the frequency of the word. The authors created the lexicon from three different sources and the resulting lexicon contained 32,051 words for which the sentiment (positive, neutral or negative) was known.  A dynamic programming algorithm was used in order to determine how many words should be present in a compressed sentence by evaluating several candidate compressed sentences. The score function was applied to each candidate sentence and the result was divided by the number of words in the compressed sentence. The compressed sentence with the highest value after this operator was deemed to be the final compressed sentence for a given uncompressed sentence. One Chinese data set of compressed sentences was manually created to evaluate their proposed approach. The study revealed that their proposed approach was able to outperform a traditional sentence compression method when the sentences were evaluated - in terms of the opinion and grammar - by a human. 

Che et al. \cite{Che:2015:Sentence} also proposed a text compression method for sentiment analysis. In their study, they make use of 10 features (which are determined for every word) to perform the compression. The features are grouped into basic, sentiment related, semantic and syntactic features. The basic features include the words before and after the word for which the features are being constructed for; they also include the POS tags for those words. There are two sentiment features, namely a binary field denoting if the word is a perception word and if it is a polarity word. These words were obtained from an existing lexicon. The semantic features include prefix and suffix characters and also Brown word features which helps identify that words which are written differently represent the same thing. They also include word embedding which makes use of Word2Vec \cite{mikolov:2013:Efficient} to determine if two words have a similar meaning. The syntactic feature makes use of a single feature which indicates the relationship between the words in the sentence. The sentence compression is conducted prior to the sentiment analysis.

With the limited amount of existing work in this field, we propose and investigate \textit{PARSEC} (PARts-of-Speech for sEntiment Compression), which makes use of an EA to achieve text compression with minimal loss in sentiment accuracy.

\section{Parts-of-Speech}\label{sec:pos}

POS, in languages, are defined as the primary groups of words that are grammatically similar. Common POS include, but are not limited to: adjective, verb and noun. For example, the words ``dog'', ``house'' and ``student'' are examples of nouns.

In this study, the Stanford Log-linear Part-Of-Speech Tagger \cite{Toutanova:2003:FeatureRich} was used to convert the original data sets into their corresponding POS tags. Thus, each word in a dataset was converted into a POS tag, and the length of each sentence in the original dataset was identical to that of the POS tagged dataset. For example, if the following sentence was found in the original dataset: ``this is a great product'', it would be converted into ``DT VBZ DT JJ NN'' (which corresponds to the POS equivalent) in the POS tagged dataset. 

We used the Penn Treebank tag set \cite{Marcus:1993:Building,Taylor:2003:ThePenn}. This set contains 36 tags and an additional 12 tags to denote certain characters such as comma, semi-colon and other characters. 

\section{Proposed Compression Method}\label{sec:proposed}

\subsection{Compressor}

\begin{figure}[ht]
  \centering
          \includegraphics[width=0.48\textwidth]{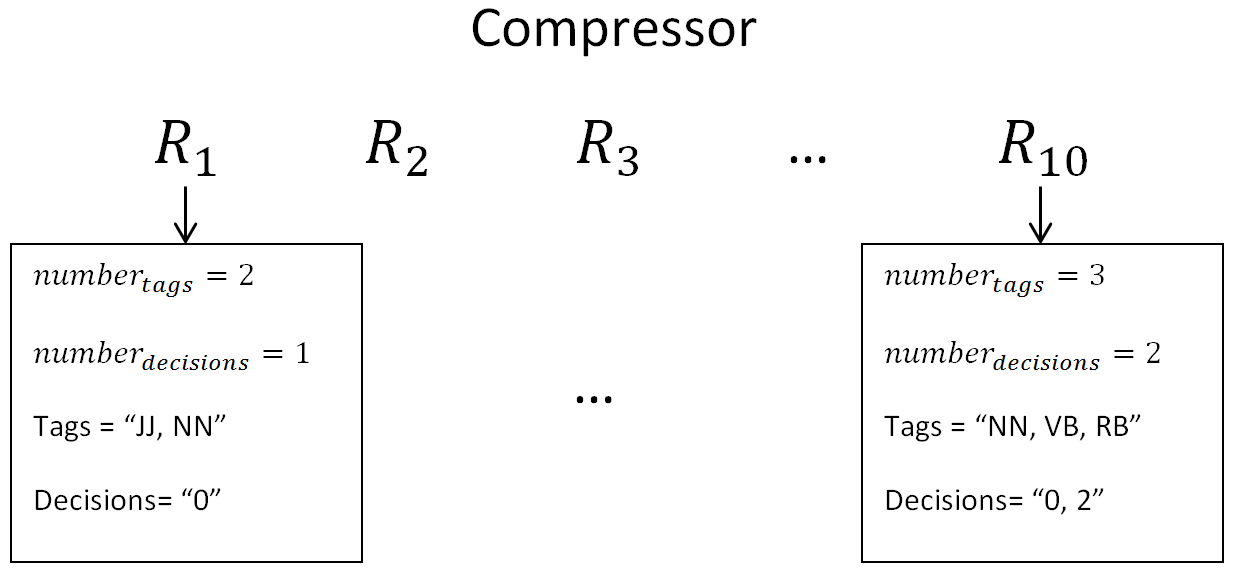}
  \caption{An example of a PARSEC compressor with 10 rules.}
\label{fig:compressor}
\end{figure}

This study proposes a new type of individual for EAs -- referred to as a \textit{compressor} -- which reduces the length of sentences when applied to a dataset. The compressor is made up of several rules, and in turn, each rule is made up of a sequence of POS tags and a decision. The decision is applied whenever it's sequence of POS tags matches the POS tags in the data. The decisions represent the indices of the words which will be removed in the matched sequence. Figure \ref{fig:compressor} illustrates an example of a compressor.

In the following explanation, let \textit{X} denote some original dataset for which \textit{PARSEC} is being applied to, and let \textit{Y} denote the equivalent POS dataset. Furthermore, let $X_{i,j}$ denote sentence $i$ and word $j$ in $X$, and $Y_{i,j}$ denote sentence $i$ and word $j$ in $Y$.

The compressor in the figure has 10 rules denoted as $R_1$, $R_2$, $R_3$, ..., $R_{10}$. The first rule, $R_1$, has 2 tags which are represented by the sequence ``JJ, NN''. This rule has one decision, and that is, to delete the word at index 0. This implies that when this particular compressor is applied, rule $R_1$ will find POS tags in $Y$ corresponding to the sequence ``JJ NN''. Let $Y_{l,m}$ and $Y_{l,m+1}$ denote two consecutive words in $Y$ which have the sequence ``JJ NN''. The decision in $R_1$ is to delete index 0, thus $R_1$ will delete $Y_{l,m}$ and consequently it will delete $X_{l,m}$. Thus, the size of $X$ and $Y$ has been reduced by one. Each time $R_1$ finds the exact sequence ``JJ NN'', the POS tag and word corresponding to ``JJ'' (index 0) in $Y$ and $X$ are deleted. 

Similarly for $R_{10}$, Let $Y_{l,m}$, $Y_{l,m+1}$ and $Y_{l,m+2}$ denote three consecutive words in $Y$ which have the sequence ``NN VB RB''. The decision in $R_{10}$ is to delete index 0 and 2, thus $R_{10}$ will delete $Y_{l,m}$, $Y_{l,m+2}$, $X_{l,m}$ and $X_{l,m+2}$ from $Y$ and $X$ respectively for each occurrence of the POS tags ``NN VB RB''.

When a rule is being processed, if a punctuation mark occurs, then the rule evaluation immediately stops. This mechanism was incorporated so that the sequence of tags -- which is encapsulated by a rule -- is applied to a single sentence and is not applied across two separate sentences. For example, consider the two following sentences ``This book was really funny. Great, tomorrow is Friday.'' which corresponds to: ``DT NN VBD RB JJ. JJ, NN VBZ NNP.'' Assume that some rule has the following tags ``JJ JJ'' and the decision is to delete both indices. If the punctuation marks are not taken into consideration, then the rule will match with the words ``funny'' and ``great''. This will result in ``This book was really . , tomorrow is Friday.'' Consequently, the sentiments in these two sentences are no longer positive. However, if punctuation is taken into consideration, then the rule will not delete any words since the full stop separates the two sentences. 

In this study, we define the term `wildcard', which represents the notion of `any POS' tag and is denoted with a `*' symbol. Thus, when a rule is being evaluated, the wildcard POS can match with any POS in the POS dataset. For example, if a rule has the following tags: ``JJ * NNP'', then any POS sequence that has `JJ', followed by any POS, followed by `NNP' will cause the rule to match with that POS sequence. 

Algorithm \ref{algo:compress} presents the generalized pseudocode to match the tags in a rule with the POS tags in the POS dataset. The algorithm also presents how to delete words from the original and POS dataset once a rule matches with a sequence.

\begin{algorithm}

\SetKwData{original}{original\_sentence}
\SetKwData{pos}{POS\_sentence}
\SetKwData{lensen}{length\_sentence}
\SetKwData{Match}{match}
\SetKwData{indices}{indices\_delete}
\SetKwData{rule}{rule}
\SetKwData{compressor}{compressor}
\SetKwData{word}{POS}
\SetKwData{tag}{tag\_in\_rule}
\SetKwData{decisions}{rule\_decisions}
\SetKwData{compressed}{compressed\_sentence}

\SetKwInOut{Input}{input}

\Input{\compressor the compressor to evaluate}
\Input{\original the original sentence}
\Input{\pos the POS sentence}
\Input{\lensen the length of the sentence to evaluate}

\Begin{

\For{each \rule in \compressor}{

	\tag $\leftarrow$ the tags inside \rule

	\decisions $\leftarrow$ the decisions inside \rule

	\For{$i\leftarrow 0$ \KwTo \lensen}{
	
		\Match $\leftarrow$ true
	
		\For{$j\leftarrow 0$ \KwTo length(\tag)}{
		
			\If{\tag[$j$] not equal \pos[$i+j$]}{
				\Match $\leftarrow$ false

				break loop and go to line 5

			}
		}

		\If{\Match equal true}{

			\For{$k\leftarrow 0$ \KwTo length(\decisions)}{

				flag \original[$i+\decisions[k]$]

			}

		}
	
	}

\compressed $\leftarrow$ delete all words in \original that have been flagged.

}

}

\Return{\compressed}
\caption{Pseudocode to apply a compressor to a sentence and compress the text.} 
\label{algo:compress} 
\end{algorithm}

\subsection{PARSEC initial population generation}

The pseudocode for creating an initial population of compressors is presented in algorithm \ref{algo:initialPopulation}. Several individuals are created based on a predefined user parameter, namely the population size. Several rules have to be created for each compressor. There must be at least one rule, and a user-defined parameter, namely max\textsubscript{rules}, is implemented to restrict the total number of rules. When creating the rules, the algorithm iterates until it has reached max\textsubscript{rules}, and a probability value of 0.5 was set to determine if a rule was to be created or not, as can be seen on line 9 in algorithm \ref{algo:initialPopulation}. Each rule consists of two primary parts, namely the tags and the decisions, the pseudocode for the creation of those respective parts are presented in algorithms \ref{algo:createPOSTags} and \ref{algo:createDecisions}.

\SetKwProg{Fn}{Function}{}{}
\begin{algorithm}

\SetKwData{popsize}{population\_size}
\SetKwData{max}{max\_tags}
\SetKwData{new}{new\_rule}
\SetKwData{newC}{new\_compressor}
\SetKwInOut{Input}{input}

\Input{\popsize the number of compressors to create}

\Begin{

	\For{$i\leftarrow 0$ \KwTo \popsize}{
	
		\textit{CreateCompressor()}

		Evaluate compressor.
	
		Add compressor to the initial population.
	}

\Fn{CreateCompressor ()}{

\newC $\leftarrow$ create a blank compressor

	\For{$j\leftarrow 0$ \KwTo \max}{
	
		\If{rand() $<$ 0.5}{

			\textit{CreateRule()} and add rule to \newC

		}
	
	}
	
	\If{Compressor  has no rule}{
		
		\textit{CreateRule()} and add rule to \newC

	}

\Return{new\_compressor}

}

\Fn{CreateRule ()}{

\new $\leftarrow$ create a blank rule

	Create the tags using algorithm \ref{algo:createPOSTags} and add the rules to \new.

	Create the decisions using algorithm \ref{algo:createDecisions} and add the rules to \new.

}

}

\Return{The initial population.}
\caption{Pseudocode to create the initial population of compressors.} 
\label{algo:initialPopulation} 
\end{algorithm}

\begin{algorithm}

\SetKwData{postags}{pos\_list}
\SetKwData{mintag}{minimum\_tags}
\SetKwData{maxtag}{maximum\_tags}
\SetKwData{numtags}{number\_tags}
\SetKwInOut{Input}{input}
\Input{\mintag the minimum number of allowed pos tags}
\Input{\maxtag the maximum number of allowed pos tags}

\Begin{
	
\postags $\leftarrow$ \{ \}

\numtags $\leftarrow$ random[\mintag, \maxtag]

	\For{$i\leftarrow 0$ \KwTo \numtags}{

		\If{$i$ = 0}{
		
			Create a tag (excluding wildcard) and add it to \postags
		
		}
		\ElseIf{$i$ = \numtags -1}{
		
			Create a tag (excluding wildcard) and add it to \postags		

		}
		\Else{
		
			Create a tag (including wildcard) and add it to \postags	
		
		}

	}

}

\Return{\postags}
\caption{Pseudocode to create POS tags.} 
\label{algo:createPOSTags} 
\end{algorithm}

\begin{algorithm}

\SetKwData{decisionlist}{decision\_list}
\SetKwData{numtags}{number\_tags}
\SetKwInOut{Input}{input}
\Input{\numtags the number of tags}

\Begin{
	
	\For{$i\leftarrow 0$ \KwTo \numtags}{
	
		\If{random[0,1] $<$ 0.5}{
		
			Add $i$ to \decisionlist
		
		}
		
	}
	
}

\Return{\decisionlist.}
\caption{Pseudocode to create decisions.} 
\label{algo:createDecisions} 
\end{algorithm}

\begin{algorithm}

\SetKwData{Sen}{sentence}
\SetKwData{Dictionary}{dictionary}
\SetKwData{NegationWords}{negation\_words}
\SetKwData{Sentiment}{sentiment\_score}
\SetKwData{Word}{word}
\SetKwData{Neg}{negation}

\SetKwInOut{Input}{input}
\SetKwInOut{Output}{output}

\Input{\Sen the sentence to be evaluated}
\Input{\Dictionary the dictionary of sentiment words}
\Input{\NegationWords the list of negation words}
\Output{The sentiment for the evaluated sentence}

\Begin{

\Sentiment $\leftarrow$ 0

\Neg $\leftarrow$ false

\For{each \Word in the \Sen}{

	\If{\Word is in \Dictionary and \Neg is true}{
		\Sentiment $\leftarrow$ \Sentiment + (-1 $\times$ \Word's sentiment value)
	}

	\If{\Word is in \Dictionary and \Neg is false}{

		\Sentiment $\leftarrow$ \Sentiment + \Word's sentiment value

	}
	
	\If{\Word is in \NegationWords}{

		\Neg $\leftarrow$ swap polarity

	}\Else{
		\Neg $\leftarrow$ false
	}
	
}

\Return{\Sentiment}

}
\caption{Pseudocode for baseline evaluation.} 
\label{algo:baseline-eval} 
\end{algorithm}

\subsection{PARSEC fitness function} \label{subsec:fitnessfunction}

The pseudocode for the fitness evaluation of a compressor is presented in algorithm \ref{algo:compressorfitness}. Each instance in the dataset is considered in turn. The sentiment for each instance in the original dataset is computed using algorithm \ref{algo:baseline-eval}. When a compressor is evaluated, it is applied to each instance and a compressed sentence is created. The sentiment of each compressed sentence is computed using algorithm \ref{algo:baseline-eval}. The compressed sentiment is then compared to the original sentiment. If these are the same then the compressed sentence has had no effect on the sentiment when compared to the original. Conversely, if the values are different, then there are two possibilities. Either the compressed sentiment is equal to the correct sentiment as labelled in the training data (in which case the compressor is rewarded) or it is not (in which case the compressor is penalised). Thus, the fitness function takes into consideration the number of new instances for which the compressor produces the correct result.

A multi-objective fitness function was created to combine the fitness, the average reduction in terms of the length of the sentences before and after compression, and the size of the compressor in terms of the number of rules. Thus, the objective was to maximize the fitness and average reduction in sentences, and to minimize the size of the compressor.

\begin{algorithm}

\SetKwData{Sens}{sentences}
\SetKwData{Sen}{sentence}
\SetKwData{OriginalLength}{original\_lengths}
\SetKwData{Fitness}{raw\_fitness}
\SetKwData{CompressedText}{compressed\_sentence}
\SetKwData{CompressedSent}{compressed\_sentiment}
\SetKwData{Compressor}{compressor}
\SetKwData{Numsens}{number\_of\_sentences}
\SetKwData{Original}{original\_sentiments}
\SetKwData{Correct}{correct\_sentiments}
\SetKwData{ChangeSize}{average\_change}
\SetKwData{NumRules}{number\_rules}

\SetKwInOut{Input}{input}
\SetKwInOut{Output}{output}

\Input{\Compressor the compressor to evaluated}
\Input{\Sen a list of sentences for which \Compressor will be evaluated on}
\Input{\OriginalLength a list of lengths for each sentence in the original data set}
\Input{\Original a list of sententiment values determined by applying algorithm \ref{algo:baseline-eval} on the original data set}
\Input{\Correct a list of correct sententiment values for the original data set}
\Input{\NumRules the number of rules that \Compressor has}
\Output{The sentiment for the evaluated sentence}

\Begin{

	\For{$i\leftarrow 1$ \KwTo \Numsens}{

		\CompressedText $\leftarrow$ apply compression to \Sens [$i$].
		
		\CompressedSent $\leftarrow$ apply algorithm \ref{algo:baseline-eval}.

		\If{\CompressedSent is not equal to \Original [$i$]}{
			
			\If{\CompressedSent is equal to \Correct [$i$]}{
			
				\Fitness $\leftarrow$ \Fitness + 1
			
			}\Else{
			
				\Fitness $\leftarrow$ \Fitness - 1
			
			}
		
		}

	\ChangeSize $\leftarrow$ \ChangeSize + (\textit{Length}(\Sens [$i$]) - \textit{Length}(\CompressedText ))

	}

\Fitness $\leftarrow$ \Fitness + [0.5 $\times$ (\ChangeSize / \Numsens)] - (0.1 $\times$ \NumRules)
	
}

\Return{\Fitness}

\caption{Pseudocode to compute fitness of a compressor.} 
\label{algo:compressorfitness} 
\end{algorithm}

\section{Experimental Setup}

\subsection{Datasets and Setup}

Twelve data sets were created by randomly selecting reviews from corresponding Amazon review datasets \cite{stanfordSnap}. For each created data set, 1000 positive and 1000 negative reviews were randomly selected from the larger corresponding Amazon data set. The larger Amazon data sets were obtained from \cite{stanfordSnap,McAuley:2015:Inferring}. The data sets were: Amazon Instant Video (AIV), Apps for Android, Automotive, Baby, Beauty, Digital Music, Health and Personal Care, Musical Instruments (MI), Patio Lawn and Garden, Pet Supplies, Tools and Home Improvement, and Toys and Games.

We propose a set of experiments whereby the compression rate is a user parameter and \textit{PARSEC} must generate compressors that can compress the original datasets to reach the specified compression rate. 

To achieve this, a lower and upper user-defined compression bound were defined. The lower compression bound (LCB) was implemented to ensure that the compressors did not result in a compression ratio less than the specified value. Similarly, the upper compression bound (UCB) ensured that the compressors did not result in a compression ratio greater than the specified value. Thus, every compressor in the EA population had to have a compression rate between the LCB and UCB. 

Furthermore, two additional user-defined parameters were implemented to ensure that the total number of rules within each compressor remained within a certain bound.  The two parameters were named $compressionRules_{min}$ and $compressionRules_{max}$. The $compressionRules_{max}$ parameter prevents compressors from having an extremely large number of rules. Both parameters aid in restricting the search space in terms of the number of rules to create and maintain for each compressor during the evolutionary process. The initial population generation had to respect the LCB and UCB, and respect the minimum and maximum number of rules. The mutation operator also had to respect these constraints.

\subsection{Sentiment analysis algorithms} \label{sec:sentimentAlgo}

Several sentiment analysis algorithms were applied to the original and compressed datasets to determine the difference in accuracy and to investigate the performance of \textit{PARSEC}. The following algorithms were used: SentiStrength \cite{Thelwall:2010:Sentiment,Thelwall:2012:Sentiment}, MeaningCloud \cite{meaningCloud}, Vivek \cite{Narayanan:2013:Fast}, Stanford \cite{Socher:2013:Recursive}, uClassify \cite{uClassify}, Sentiment140 \cite{sent140}, Intellexer \cite{intellexer} and LingPipe \cite{lingpipe}.

\subsection{Execution of experiments}

We first applied the sentiment analysis algorithms described in section \ref{sec:sentimentAlgo} to the original datasets and record the test accuracy. Then, we ran \textit{PARSEC} using the baseline evaluation to create compressor rules. Once the compressors were created, we applied the compressors to the original datasets to generate the compressed data. We applied the sentiment analysis algorithms on the compressed data and recorded the test accuracy. In the following section we report on the change in accuracy between the compressed and original data. \textit{PARSEC} evolved a population of 250 compressors over 100 generations, the crossover rate was 60\% and mutation rate was 40\%.

\begin{table}
\protect\caption{The lower (LCB) and upper (UCB) compression bound constrain the algorithm by ensuring that the compression rate for each compressor is between the two values respectively. The table presents the LCB and UCB values (\%) along with the minimum and maximum number of rules used in the experiments. These set compression rates were enforced and the performance of \textit{PARSEC} was measured on several datasets. The values for the minimum and maximum number of rules were determined by additional trial runs. It was observed that a greater number of rules were needed to achieve higher compression rates.} \label{table:ratesAndRules}

\begin{centering}
\begin{tabular}{|c|c|p{1.6cm}|p{1.6cm}|}
\hline
\textbf{LCB} & \textbf{UCB} & \textbf{$compression
\allowbreak
Rules_{min}$} & \textbf{$compression
\allowbreak
Rules_{max}$}\tabularnewline
\hline 
10 & 13 & 5 & 50\tabularnewline
15 & 18 & 10 & 70\tabularnewline
20 & 13 & 20 & 90\tabularnewline
25 & 28 & 40 & 120\tabularnewline
30 & 33 & 90 & 150\tabularnewline
50 & 53 & 350 & 500\tabularnewline
\hline
\end{tabular}
\par\end{centering}

\end{table}

\section{Results and Conclusion}

\begin{figure}[!h]
  \centering
          \includegraphics[width=0.45\textwidth]{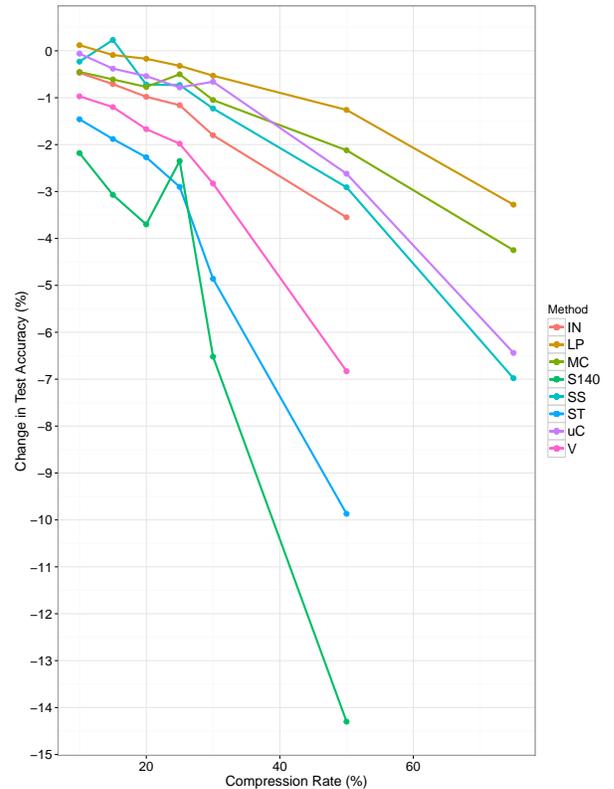}
  \caption{Change in average test accuracy across the different sentiment analysis algorithms for different text compression rates enforced by PARSEC, averaged across all the data sets. A higher positive value indicates a better result. The various sentiment analysis technologies are denoted as follows: \emph{IN, LP, MC}, \emph{S140},
\emph{SS}, \emph{ST}, \emph{uC} and \emph{V} refer to Intellexer, LingPipe, MeaningCloud, Sentiment140, SentiStrength, Stanford, uClassify and Vivek respectively. The accuracy obtained by LingPipe outperformed all the other methods. Furthermore, LingPipe obtained an average positive change in accuracy of 0.1\% for 10\% compression. At a compression rate of 50\%, LingPipe obtained an change in accuracy of -1.2\% averaged over all the datasets, illustrating that it works well with PARSEC.}
\label{fig:changeTrend}
\end{figure}

Figure \ref{fig:changeTrend} shows the average change in test accuracy across the different methods based on the compression rates. The general trend is a decrease in performance with the largest decrease in performance  observed by Sentiment140. LingPipe and SentiStrength were the only two methods that achieved a positive change in performance (on certain datasets) on compression rates of 10\% and 15\% respectively. When experimenting with \textit{PARSEC} we posed the following question. For a predetermined drop in accuracy, how much compression is achievable by \textit{PARSEC}? For a threshold of -1\% compression, Vivek and LingPipe were able to achieve up to 30\% compression. For this same threshold five methods could achieve up to 20\% compression rate. Regardless of the compression rate used, both Stanford and Sentiment140 did not obtain any result less than -1\%. These two algorithms produced the weakest performance with \textit{PARSEC}.

For a threshold of -2\%, five algorithms could achieve up to 30\% compression rate and LingPipe was able to achieve up to 50\% compression. When the threshold is set to -3\% four methods could achieve up to 50\% compression. 

We additionally wanted to determine the change in test accuracy when a much larger compression rate was used. Table \ref{table:75Rate} presents the results when a compression rate of 75\% was used. Only certain algorithms were tested due to our limited access to the sentiment analysis algorithms. These results reveal that a large compression rate is achievable without a great loss in test accuracy: LingPipe obtained a loss of -3.3\% accuracy.

An example of \textit{PARSEC} compression is as follows. The original sentence (which contains 56 words and is already stemmed): {\em ``this be a wonderful movie involve fascinating people who be such great actor and actress. the event be interesting and I never get tire of watch the episode. I be so sad to see that they be not go to keep go with the life of all these inhabitant of Lark rise and candleford. Mrs. Hamlin''}. The compressed sentence (22 words): {\em ``this be wonderful movie involve fascinating people who actress. get tire. so sad to see not keep go rise candleford. Mrs. Hamlin''}.

\begin{table}[!h]
\protect\caption{Average change in test accuracy (\%) across all the data sets when
PARSEC models were created for 75\% compression rate. } \label{table:75Rate}
\begin{centering}
\begin{tabular}{|l|c|c|c|}
\hline 
\textbf{Algorithm} & \vtop{\hbox{\strut \textbf{Change in}}\hbox{\strut \textbf{Test Accuracy}}} & \vtop{\hbox{\strut \textbf{Original}}\hbox{\strut \textbf{Accuracy}}}& \vtop{\hbox{\strut \textbf{Compressed}}\hbox{\strut \textbf{Accuracy}}} \tabularnewline
\hline 
LingPipe & -3.2 & 79.7 & 76.4\tabularnewline
MeaningCloud & -4.2 & 62.5 & 58.2\tabularnewline
SentiStrength & -6.9 & 67.8 & 60.8\tabularnewline
uClassify & -6.4 & 80.7 & 74.3\tabularnewline
\hline 
\end{tabular}
\par\end{centering}

\end{table}

Below are four rules which were randomly selected from a \textit{PARSEC} model to illustrate some of the evolved rules. This model had a total of 38 rules. The POS tags are presented in order of appearance between square brackets. The indices start from zero. Rule 1 denotes that if the POS  ``NNP'' is followed by ``JJ'', then delete both words. Rule 4 denotes that if ``IN'' is followed by any POS and then followed by ``VB'', then delete the word at index 1. 

\begin{itemize}
\item Rule 1: [NNP, JJ]. Number of decisions: 2. \\Decisions: [0, 1]
\item Rule 2: [VBG, NNP]. Number of decisions: 2. \\Decisions: [0, 1]
\item Rule 3: [IN, CC]. Number of decisions: 2. \\Decisions: [0, 1]
\item Rule 4: [IN, *, VB]. Number of decisions: 1. \\Decisions: [1]
\end{itemize}

This study empirically demonstrates that machine learning can reduce the amount of data needed to determine the sentiments within sentences with little loss in accuracy. To achieve this, we proposed an evolutionary algorithm, \textit{PARSEC}, which evolves a population of chromosomes which encode rules for compression based on the removal of POS. 

We studied the test accuracies achieved by the eight algorithms when set compression rate thresholds were enforced. Two algorithms showed marginal improvements in accuracy on several data sets for a compression rate of 15\%. Additionally, there were 9 data sets whereby some of the algorithms were able to yield an improvement in accuracy for a compression rate of 30\%. At a threshold of -1\% sentiment accuracy loss, two methods were able to achieve up to 30\% compression, five methods up to 20\% compression, and six methods up to 10\% compression. The best performing algorithm, LingPipe, showed only modest losses in accuracy even at high compression: 3.3\% for a four-fold data compression. 

It would be of interest to incorporate the \textit{PARSEC} algorithm directly into a sentiment analysis algorithm instead of via a simple dictionary approach. One could achieve this by replacing the fitness function used in the evolutionary phase with a more sophisticated sentiment analysis algorithm. This would allow the emergence of an algorithm where the compression and sentiment analysis are mutually optimised for best performance. 

\section*{Acknowledgment}
The financial assistance of the National Research Foundation (NRF) towards this research is hereby acknowledged. Opinions expressed and conclusions arrived at, are those of the authors and are not necessarily to be attributed to the NRF.

\bibliographystyle{IEEEtran}

\begin{thebibliography}{10}
\providecommand{\url}[1]{#1}
\csname url@samestyle\endcsname
\providecommand{\newblock}{\relax}
\providecommand{\bibinfo}[2]{#2}
\providecommand{\BIBentrySTDinterwordspacing}{\spaceskip=0pt\relax}
\providecommand{\BIBentryALTinterwordstretchfactor}{4}
\providecommand{\BIBentryALTinterwordspacing}{\spaceskip=\fontdimen2\font plus
\BIBentryALTinterwordstretchfactor\fontdimen3\font minus
  \fontdimen4\font\relax}
\providecommand{\BIBforeignlanguage}[2]{{%
\expandafter\ifx\csname l@#1\endcsname\relax
\typeout{** WARNING: IEEEtran.bst: No hyphenation pattern has been}%
\typeout{** loaded for the language `#1'. Using the pattern for}%
\typeout{** the default language instead.}%
\else
\language=\csname l@#1\endcsname
\fi
#2}}
\providecommand{\BIBdecl}{\relax}
\BIBdecl

\bibitem{Pang:2008:Opinion}
B.~Pang and L.~Lee, ``Opinion mining and sentiment analysis,'' \emph{Found.
  Trends Inf. Retr.}, vol.~2, no. 1-2, pp. 1--135, Jan. 2008.

\bibitem{liu:2015:sentiment}
B.~Liu, \emph{Sentiment Analysis: Mining Opinions, Sentiments, and
  Emotions}.\hskip 1em plus 0.5em minus 0.4em\relax Cambridge University Press,
  2015.

\bibitem{Ravi:2015:ASurvey}
K.~Ravi and V.~Ravi, ``A survey on opinion mining and sentiment analysis,''
  \emph{Know.-Based Syst.}, vol.~89, no.~C, pp. 14--46, Nov. 2015.

\bibitem{Eiben:2003:IntroTo}
A.~E. Eiben and J.~E. Smith, \emph{Introduction to Evolutionary
  Computing}.\hskip 1em plus 0.5em minus 0.4em\relax SpringerVerlag, 2003.

\bibitem{Yu:2012:Introduction}
X.~Yu and M.~Gen, \emph{Introduction to Evolutionary Algorithms}.\hskip 1em
  plus 0.5em minus 0.4em\relax Springer Publishing Company, Incorporated, 2012.

\bibitem{Feng:2010:AChinese}
S.~Feng, D.~Wang, G.~Yu, B.~Li, and K.-F. Wong, \emph{A Chinese Sentence
  Compression Method for Opinion Mining}.\hskip 1em plus 0.5em minus
  0.4em\relax Berlin, Heidelberg: Springer Berlin Heidelberg, 2010, pp.
  320--329.

\bibitem{Che:2015:Sentence}
W.~Che, Y.~Zhao, H.~Guo, Z.~Su, and T.~Liu, ``Sentence compression for
  aspect-based sentiment analysis,'' \emph{IEEE/ACM Trans. Audio, Speech and
  Lang. Proc.}, vol.~23, no.~12, pp. 2111--2124, Dec. 2015.

\bibitem{mikolov:2013:Efficient}
T.~Mikolov, K.~Chen, G.~Corrado, and J.~Dean, ``Efficient estimation of word
  representations in vector space,'' \emph{arXiv preprint arXiv:1301.3781},
  2013.

\bibitem{Toutanova:2003:FeatureRich}
K.~Toutanova, D.~Klein, C.~D. Manning, and Y.~Singer, ``Feature-rich
  part-of-speech tagging with a cyclic dependency network,'' in
  \emph{Proceedings of the 2003 Conference of the North American Chapter of the
  Association for Computational Linguistics on Human Language Technology -
  Volume 1}, ser. NAACL '03.\hskip 1em plus 0.5em minus 0.4em\relax
  Stroudsburg, PA, USA: Association for Computational Linguistics, 2003, pp.
  173--180.

\bibitem{Marcus:1993:Building}
M.~P. Marcus, M.~A. Marcinkiewicz, and B.~Santorini, ``Building a large
  annotated corpus of english: The penn treebank,'' \emph{Comput. Linguist.},
  vol.~19, no.~2, pp. 313--330, Jun. 1993.

\bibitem{Taylor:2003:ThePenn}
A.~Taylor, M.~Marcus, and B.~Santorini, \emph{Treebanks: Building and Using
  Parsed Corpora}.\hskip 1em plus 0.5em minus 0.4em\relax Dordrecht: Springer
  Netherlands, 2003, ch. The Penn Treebank: An Overview, pp. 5--22.

\bibitem{stanfordSnap}
J.~Leskovec and A.~Krevl, ``{SNAP Datasets}: {Stanford} large network dataset
  collection,'' \url{http://snap.stanford.edu/data}, Jun. 2014.

\bibitem{McAuley:2015:Inferring}
J.~McAuley, R.~Pandey, and J.~Leskovec, ``Inferring networks of substitutable
  and complementary products,'' in \emph{Proceedings of the 21th ACM SIGKDD
  International Conference on Knowledge Discovery and Data Mining}, ser. KDD
  '15.\hskip 1em plus 0.5em minus 0.4em\relax New York, NY, USA: ACM, 2015, pp.
  785--794.

\bibitem{Thelwall:2010:Sentiment}
M.~Thelwall, K.~Buckley, G.~Paltoglou, D.~Cai, and A.~Kappas, ``Sentiment in
  short strength detection informal text,'' \emph{J. Am. Soc. Inf. Sci.
  Technol.}, vol.~61, no.~12, pp. 2544--2558, Dec. 2010.

\bibitem{Thelwall:2012:Sentiment}
M.~Thelwall, K.~Buckley, and G.~Paltoglou, ``Sentiment strength detection for
  the social web,'' \emph{J. Am. Soc. Inf. Sci. Technol.}, vol.~63, no.~1, pp.
  163--173, Jan. 2012.

\bibitem{meaningCloud}
MeaningCloud, ``Meaningcloud is a brand by meaningcloud llc, a s|ngular
  company.'' \url{https://www.meaningcloud.com/}, 2015.

\bibitem{Narayanan:2013:Fast}
V.~Narayanan, I.~Arora, and A.~Bhatia, ``Fast and accurate sentiment
  classification using an enhanced naive bayes model,'' in \emph{Proceedings of
  the 14th International Conference on Intelligent Data Engineering and
  Automated Learning --- IDEAL 2013 - Volume 8206}, ser. IDEAL 2013.\hskip 1em
  plus 0.5em minus 0.4em\relax New York, NY, USA: Springer-Verlag New York,
  Inc., 2013, pp. 194--201.

\bibitem{Socher:2013:Recursive}
R.~Socher, A.~Perelygin, J.~Y. Wu, J.~Chuang, C.~D. Manning, A.~Y. Ng, and
  C.~Potts, ``Recursive deep models for semantic compositionality over a
  sentiment treebank,'' in \emph{Proceedings of the conference on empirical
  methods in natural language processing (EMNLP)}, 2013, pp. 1631--1642.

\bibitem{uClassify}
uClassify, ``uclassify,'' \url{https://www.uclassify.com/}, 2008.

\bibitem{sent140}
A.~Go, R.~Bhayani, and L.~Huang, ``Twitter sentiment classification using
  distant supervision,'' \emph{Technical report, Stanford}, 2009.

\bibitem{intellexer}
EffectiveSoft, ``Intellexer,'' \url{http://www.intellexer.com/}, 2000.

\bibitem{lingpipe}
Alias-i., ``Lingpipe 4.1.0,'' \url{http://alias-i.com/lingpipe}, 2008.

\end{thebibliography}


\appendix
\section{Detailed results for fixed compression rate experiments} \label{appendix:Results}

Tables \ref{table:compression10} to \ref{table:compression50} present the difference in test accuracy results for the various compression rates. The various sentiment analysis algorithms are abbreviated as follows: \emph{IN, LP, MC}, \emph{S140},
\emph{SS}, \emph{ST}, \emph{uC} and \emph{V} refer to Intellexer, LingPipe, MeainingCloud, Sentiment140, SentiStrength, Stanford, uClassify and Vivek respectively. For a compression rate of 10\%, the best and worst change in accuracy were obtained by LingPipe and Sentiment140 respectively. Based on all the findings for 10 per cent compression, there were 21 cases across all the methods and data sets whereby the change in accuracy was greater than zero. For a compression rate of 15\% the number of cases was 17 whereby LingPipe obtained the best average change in accuracy with a value of 0.0\%. 

The best change in accuracy for 20\% compression rate was obtained by uClassify with a value of 0.7 on two data sets. There were six data sets for which uClassify reduced the data up to 20\% and for which the test accuracy on the compressed data was better than on the original data. Additionally, there were 14 cases in total across all the algorithms whereby the change in accuracy was greater than zero for 20\% compression. 

The findings revealed that a for the compression rate of 25\% there were 14 cases for which the change was greater than zero, and the best change in average accuracy was obtained once again by LingPipe with a value of -0.3\%. There was a reduction performance in terms of the number of cases having a change greater than zero when a compression of 30\% was used; notably 9 cases. uClassify however, on 5 datasets, had a test accuracy greater than zero. Finally, only two algorithms were able to obtain improvements in terms of the changes for a compression rate of 50\%, namely uClassify and LingPipe, with changes in 3 and 2 data sets respectively. 

\newpage

\begin{table}[!h]

\begin{centering}

\begin{tabular}{|>{\centering}p{1.2cm}|>{\centering}p{0.59cm}|>{\centering}p{0.59cm}|>{\centering}p{0.59cm}|>{\centering}p{0.59cm}|>{\centering}p{0.59cm}|>{\centering}p{0.59cm}|>{\centering}p{0.59cm}|c|}
\hline
 & \textbf{IN} & \textbf{LP} & \textbf{MC} & \textbf{S140} & \textbf{SS} & \textbf{ST} & \textbf{uC} & \textbf{V}\\
\hline 
{AIV} & -0.9 & -0.0 & 0.4 & -2.1 & -0.3 & -3.0 & -1.9 & -0.9\tabularnewline\hline
{MI} & 0.1 & 0.3 & -0.7 & -1.9 & -0.4 & 0.5 & -0.4 & -1.4\tabularnewline\hline
{Digital Music} & -1.0 & 0.2 & -0.3 & -2.5 & -0.1 & -1.9 & -0.4 & -2.0\tabularnewline\hline
{Baby} & 0.0 & 0.4 & -0.2 & -2.7 & 0.0 & -1.2 & 0.3 & 0.2\tabularnewline\hline
{Patio \& Garden} & -0.1 & 0.4 & -0.2 & -2.1 & -0.2 & -0.7 & 0.9 & -0.9\tabularnewline\hline
{Automotive} & -1.1 & 0.4 & -0.5 & -2.1 & 0.0 & -1.1 & 0.7 & -0.9\tabularnewline\hline
{Pet Supplies} & -1.2 & 0.3 & -0.4 & -2.9 & -0.4 & 0.3 & 0.0 & -1.6\tabularnewline\hline
{Apps for Android} & -1.4 & 0.0 & -0.5 & -2.5 & -0.3 & -3.5 & 0.0 & -1.6\tabularnewline\hline
{Beauty} & -0.5 & 0.0 & 0.2 & -1.2 & 0.0 & -0.5 & -0.7 & -0.4\tabularnewline\hline
{Tools \& Home} & 0.8 & -0.1 & -1.0 & -2.3 & -0.6 & -1.7 & 0.5 & 0.0\tabularnewline\hline
{Toys \& Games} & -0.5 & 0.0 & -1.5 & -1.8 & 0.2 & -3.1 & -0.1 & -1.4\tabularnewline\hline
{Health \& Personal} & 0.4 & 0.1 & -0.4 & -1.6 & -0.3 & -1.4 & 0.5 & -0.4\tabularnewline\hline
\textbf{Average} & -0.4  & 0.1  & -0.4  & -2.1  & -0.2  & -1.4  & 0.0  & -0.9 \tabularnewline
\hline
\end{tabular}
\par\end{centering}

\protect\caption{Difference between the compressed and original test accuracy when a compression rate between 10\% and 13\% was imposed.} \label{table:compression10}
\end{table}

\begin{table} [!h]
\begin{centering}
\begin{tabular}{|>{\centering}p{1.2cm}|>{\centering}p{0.59cm}|>{\centering}p{0.59cm}|>{\centering}p{0.59cm}|>{\centering}p{0.59cm}|>{\centering}p{0.59cm}|>{\centering}p{0.59cm}|>{\centering}p{0.59cm}|c|}
\hline
 & \textbf{IN} & \textbf{LP} & \textbf{MC} & \textbf{S140} & \textbf{SS} & \textbf{ST} & \textbf{uC} & \textbf{V}\\
\hline 
{AIV} & -1.1 & -1.1 & -0.2 & -2.7 & 0.1 & -3.2 & -2.0 & -0.6\tabularnewline\hline
{MI} & 1.1 & 0.5 & -0.8 & -2.9 & -0.8 & 0.0 & -1.3 & -0.8\tabularnewline\hline
{Digital Music} & -1.6 & 0.4 & -0.7 & -3.6 & -0.5 & -2.3 & -1.2 & -2.1\tabularnewline\hline
{Baby} & 0.0 & -0.9 & -0.5 & -2.4 & 0.2 & -2.0 & 0.6 & -0.1\tabularnewline\hline
{Patio \& Garden} & -0.3 & -0.4 & -0.3 & -3.5 & -0.6 & -1.2 & 0.5 & -1.0\tabularnewline\hline
{Automotive} & -1.4 & 0.6 & -1.0 & -2.8 & -0.3 & -2.2 & 0.9 & -0.8\tabularnewline\hline
{Pet Supplies} & -1.1 & 0.6 & -0.2 & -3.8 & -0.6 & -1.3 & 0.2 & -2.9\tabularnewline\hline
{Apps for Android} & -1.7 & -0.3 & -0.9 & -4.2 & -0.5 & -3.4 & -1.0 & -3.2\tabularnewline\hline
{Beauty} & -0.9 & -0.3 & -0.7 & -3.5 & -0.2 & -1.3 & -1.8 & -1.2\tabularnewline\hline
{Tools \& Home} & -0.2 & 0.0 & 0.0 & -3.0 & -0.3 & -1.7 & 1.2 & 0.7\tabularnewline\hline
{Toys \& Games} & -1.3 & -0.2 & -1.2 & -1.6 & -0.1 & -1.1 & -0.8 & -1.8\tabularnewline\hline
{Health \& Personal} & 0.1 & 0.0 & -0.6 & -2.8 & -0.5 & -2.6 & 0.2 & -0.4\tabularnewline\hline
\textbf{Average} & -0.7  & 0.0  & -0.6  & -3.0  & -0.3  & -1.8  & -0.3  & -1.2 \tabularnewline
\hline
\end{tabular}
\par\end{centering}

\protect\caption{Difference between the compressed and original test accuracy when a compression rate between 15\% and 18\% was imposed. } \label{table:compression15}
\end{table}

\begin{table*}
\begin{centering}
\begin{tabular}{|l|c|c|c|c|c|c|c|c|}
\hline 
 & \textbf{IN} & \textbf{LP} & \textbf{MC} & \textbf{S140} & \textbf{SS} & \textbf{ST} & \textbf{uC} & \textbf{V}\tabularnewline
\hline 
{AIV} & -1.7 & -1.2 & 0.3 & -4.0 & -0.4 & -3.1 & -1.8 & -1.8\tabularnewline\hline 
{MI} & 0.2 & 1.1 & -1.1 & -2.5 & -0.6 & 0.4 & -1.2 & -1.4\tabularnewline\hline 
{Digital Music} & -2.2 & -0.4 & -0.2 & -3.5 & -1.2 & -3.1 & -1.7 & -2.0\tabularnewline\hline 
{Baby} & -0.7 & -0.4 & -0.4 & -2.7 & -0.8 & -2.2 & 0.5 & -1.1\tabularnewline\hline 
{Patio \& Garden} & -0.9 & -1.1 & -1.2 & -3.7 & -0.5 & -2.2 & 0.5 & -0.9\tabularnewline\hline 
{Automotive} & -1.6 & 0.2 & -0.3 & -3.5 & -0.9 & -2.8 & 0.3 & -1.7\tabularnewline\hline 
{Pet Supplies} & -1.0 & 0.4 & -1.3 & -3.5 & -0.4 & -2.1 & 0.7 & -2.2\tabularnewline\hline 
{Apps for Android} & -1.7 & -0.1 & -1.6 & -4.8 & -0.4 & -4.5 & -1.7 & -3.3\tabularnewline\hline 
{Beauty} & -1.1 & 0.2 & -0.6 & -4.0 & -1.0 & -1.1 & -1.2 & -1.2\tabularnewline\hline 
{Tools \& Home} & 0.6 & -0.1 & -1.2 & -4.4 & -0.7 & -1.9 & 0.0 & -1.5\tabularnewline\hline 
{Toys \& Games} & -1.5 & 0.0 & -0.4 & -3.3 & -0.2 & -2.8 & -1.5 & -1.9\tabularnewline\hline 
{Health \& Personal} & 0.0 & -0.4 & -1.0 & -4.2 & -1.2 & -1.7 & 0.7 & -0.9\tabularnewline\hline 
\textbf{Average} & -0.9  & -0.1  & -0.7  & -3.7  & -0.7  & -2.2  & -0.5  & -1.6 \tabularnewline
\hline
\end{tabular}
\par\end{centering}

\protect\caption{Difference between the compressed and original test accuracy
when a compression rate between 20\% and 23\% was imposed.} \label{table:compression20}
\end{table*}

\begin{table*}
\begin{centering}
\begin{tabular}{|l|c|c|c|c|c|c|c|c|}
\hline 
 & \textbf{IN} & \textbf{LP} & \textbf{MC} & \textbf{S140} & \textbf{SS} & \textbf{ST} & \textbf{uC} & \textbf{V}\tabularnewline
\hline 
{AIV} & -2.4 & -1.6 & -0.6 & -3.7 & -0.2 & -6.0 & -2.5 & -2.0\tabularnewline\hline 
{MI} & -0.6 & 1.0 & -0.1 & -4.8 & -1.0 & -1.6 & -0.1 & -1.8\tabularnewline\hline 
{Digital Music} & -1.9 & -0.1 & -0.5 & 25.1 & -0.5 & -2.4 & -2.9 & -3.5\tabularnewline\hline 
{Baby} & 0.1 & -0.9 & 0.5 & -4.4 & -0.5 & -2.6 & -0.4 & -0.6\tabularnewline\hline 
{Patio \& Garden} & -0.9 & -1.4 & 0.1 & -4.7 & -0.9 & -1.5 & 0.6 & -1.6\tabularnewline\hline 
{Automotive} & -2.0 & 0.1 & 0.1 & -3.3 & -1.5 & -4.2 & 0.5 & -0.4\tabularnewline\hline 
{Pet Supplies} & -0.8 & 0.0 & -0.7 & -5.3 & -0.6 & -1.7 & 0.1 & -2.9\tabularnewline\hline 
{Apps for Android} & -2.4 & -0.4 & -1.4 & -6.7 & -0.9 & -3.4 & -0.9 & -3.4\tabularnewline\hline 
{Beauty} & -1.1 & 0.2 & -0.7 & -5.5 & -0.6 & -2.1 & -1.8 & -2.4\tabularnewline\hline 
{Tools \& Home} & -0.8 & -1.0 & -1.6 & -5.2 & -1.1 & -3.6 & 0.0 & -0.3\tabularnewline\hline 
{Toys \& Games} & -1.2 & -0.5 & -1.1 & -5.7 & -0.2 & -2.7 & -1.0 & -3.6\tabularnewline\hline 
{Health \& Personal} & 0.3 & 0.9 & 0.0 & -3.9 & -0.3 & -2.8 & -0.9 & -1.1\tabularnewline\hline 
\textbf{Average} & -1.1 & -0.3  & -0.5  & -2.3  & -0.7  & -2.9  & -0.7  & -1.9 \tabularnewline
\hline 
\end{tabular}
\par\end{centering}

\protect\caption{Difference between the compressed and original test accuracy
when a compression rate between 25\% and 28\% was imposed.} \label{table:compression25}
\end{table*}

\begin{table*}
\begin{centering}
\begin{tabular}{|l|c|c|c|c|c|c|c|c|}
\hline
 & \textbf{IN} & \textbf{LP} & \textbf{MC} & \textbf{S140} & \textbf{SS} & \textbf{ST} & \textbf{uC} & \textbf{V}\tabularnewline
\hline 
{AIV} & -2.3 & -2.1 & -0.1 & -8.2 & -1.3 & -7.0 & -2.3 & -2.5\tabularnewline\hline 
{MI} & -0.5 & 0.6 & -1.1 & -6.9 & -1.6 & -2.4 & -1.9 & -2.8\tabularnewline\hline 
{Digital Music} & -3.3 & -1.1 & -1.0 & -6.2 & -1.7 & -3.6 & -2.8 & -4.9\tabularnewline\hline 
{Baby} & -0.8 & -0.7 & 0.2 & -6.1 & -1.1 & -5.3 & 0.7 & -1.3\tabularnewline\hline 
{Patio \& Garden} & -1.8 & -1.2 & -1.7 & -6.6 & -1.5 & -4.3 & -0.3 & -2.2\tabularnewline\hline 
{Automotive} & -1.2 & -0.4 & -0.9 & -4.8 & -1.5 & -4.7 & 0.8 & -2.6\tabularnewline\hline 
{Pet Supplies} & -2.2 & -0.9 & -1.1 & -5.8 & -1.0 & -4.4 & 1.6 & -2.8\tabularnewline\hline 
{Apps for Android} & -3.0 & 0.0 & -2.1 & -7.7 & -1.1 & -6.4 & -2.1 & -5.3\tabularnewline\hline 
{Beauty} & -2.4 & -0.2 & -1.1 & -7.7 & -1.2 & -4.6 & -1.9 & -3.2\tabularnewline\hline 
{Tools \& Home} & -0.9 & 0.0 & -1.3 & -5.9 & -1.5 & -5.0 & 0.4 & -1.1\tabularnewline\hline 
{Toys \& Games} & -2.2 & 0.0 & -1.6 & -6.0 & -0.2 & -5.0 & -0.7 & -3.0\tabularnewline\hline 
{Health \& Personal} & -0.6 & -0.3 & -0.60 & -6.10 & -0.70 & -5.30 & 0.70 & -2.00\tabularnewline\hline 
\textbf{Average} & -1.8 & -0.5  & -1.0  & -6.5  & -1.2  & -4.8  & -0.6  & -2.8 \tabularnewline
\hline 
\end{tabular}
\par\end{centering}

\protect\caption{Difference between the compressed and original test accuracy
when a compression rate between 30\% and 33\% was imposed.} \label{table:compression30}
\end{table*}

\begin{table*}
\begin{centering}
\begin{tabular}{|l|c|c|c|c|c|c|c|c|}
\hline
 & \textbf{IN} & \textbf{LP} & \textbf{MC} & \textbf{S140} & \textbf{SS} & \textbf{ST} & \textbf{uC} & \textbf{V}\tabularnewline
\hline 
{AIV} & -5.8 & -4.2 & -1.8 & -15.6 & -3.4 & -13.2 & -7.8 & -8.2\tabularnewline\hline 
{MI} & -1.9 & 0.8 & -0.9 & -15.1 & -2.6 & -6.9 & -3.9 & -7.6\tabularnewline\hline 
{Digital Music} & -5.5 & -1.3 & -1.9 & -15.3 & -3.5 & -9.9 & -5.5 & -9.0\tabularnewline\hline 
{Baby} & -1.4 & -1.4 & -1.7 & -12.4 & -2.4 & -11.2 & -1.5 & -5.6\tabularnewline\hline 
{Patio \& Garden} & -3.8 & -1.8 & -2.1 & -13.9 & -2.5 & -10.1 & -0.5 & -4.7\tabularnewline\hline 
{Automotive} & -3.7 & -2.4 & -1.9 & -12.6 & -2.8 & -7.9 & 0.1 & -4.6\tabularnewline\hline 
{Pet Supplies} & -1.8 & -1.6 & -2.2 & -14.9 & -2.5 & -8.0 & 0.6 & -6.2\tabularnewline\hline 
{Apps for Android} & -6.2 & -1.6 & -4.8 & -17.1 & -4.4 & -12.4 & -4.3 & -11.0\tabularnewline\hline 
{Beauty} & -3.0 & -0.1 & -1.8 & -15.3 & -3.0 & -8.7 & -4.1 & -7.8\tabularnewline\hline 
{Tools \& Home} & -2.2 & 0.1 & -2.2 & -14.3 & -3.6 & -9.5 & 1.0 & -4.9\tabularnewline\hline 
{Toys \& Games} & -4.8 & -1.2 & -2.0 & -12.7 & -1.8 & -10.1 & -4.3 & -7.2\tabularnewline\hline 
{Health \& Personal} & -2.2 & -0.2 & -2.0 & -12.1 & -2.2 & -10.1 & -1.1 & -5.0\tabularnewline\hline 
\textbf{Average} & -3.5  & -1.2  & -2.1  & -14.3  & -2.9  & -9.8  & -2.6  & -6.8 \tabularnewline
\hline 
\end{tabular}
\par\end{centering}

\protect\caption{Difference between the compressed and original test accuracy when a compression rate between 50\% and 53\% was imposed.} \label{table:compression50}
\end{table*}

\end{document}